\documentclass[sn-apa]{sn-jnl}

\usepackage{graphicx}%
\usepackage{multirow}%
\usepackage{amsmath,amssymb,amsfonts}%
\usepackage{amsthm}%
\usepackage{mathrsfs}%
\usepackage[title]{appendix}%
\usepackage{xcolor}%
\usepackage{textcomp}%
\usepackage{manyfoot}%
\usepackage{booktabs}%
\usepackage{algorithm}%
\usepackage{algorithmicx}%
\usepackage{algpseudocode}%
\usepackage{listings}%

\begin{document}

\title{Estimating the distribution of numerosity and non-numerical visual magnitudes in natural scenes using computer vision}

\author[1]{\fnm{Kuinan} \sur{Hou}}\email{kuinan.hou@phd.unipd.it}
\author[1,2]{\fnm{Marco} \sur{Zorzi}}\email{marco.zorzi@unipd.it}
\author*[1,3]{\fnm{Alberto} \sur{Testolin}}\email{alberto.testolin@unipd.it}

\affil[1]{\orgdiv{Department of General Psychology}, \orgname{University of Padova}}
\affil[2]{\orgname{IRCCS San Camillo Hospital}, \orgaddress{\city{Venice-Lido}}}
\affil[3]{\orgdiv{Department of Mathematics}, \orgname{University of Padova}}

\abstract{
Humans share with many animal species the ability to perceive and approximately represent the number of objects in visual scenes. This ability improves throughout childhood, suggesting that learning and development play a key role in shaping our number sense. This hypothesis is further supported by computational investigations based on deep learning, which have shown that numerosity perception can spontaneously emerge in neural networks that learn the statistical structure of images with a varying number of items. 
However, neural network models are usually trained using synthetic datasets that might not faithfully reflect the statistical structure of natural environments, and there is also growing interest in using more ecological visual stimuli to investigate numerosity perception in humans.
In this work, we exploit recent advances in computer vision algorithms to design and implement an original pipeline that can be used to estimate the distribution of numerosity and non-numerical magnitudes in large-scale datasets containing thousands of real images depicting objects in daily life situations. We show that in natural visual scenes the frequency of appearance of different numerosities follows a power law distribution. Moreover, we show that the correlational structure for numerosity and continuous magnitudes is stable across datasets and scene types (homogeneous vs. heterogeneous object sets). We suggest that considering such ``ecological'' pattern of covariance is important to understand the influence of non-numerical visual cues on numerosity judgements.
}


\maketitle

\section{Introduction}

Humans can approximately enumerate the number of objects in a visual scene, even without counting, and such a skill has been documented in many animal species and children prior to language development \citep{dehaene2011number}. This ability, often referred to as “number sense”, is thought to be highly adaptive for survival \citep{nieder2005counting}, and the fact that it is shared across species has led many to suggest that it could be part of our evolutionary endowment \citep{ferrigno2017evolutionary}.

At the same time, number sense gets progressively refined throughout childhood \citep{piazza2010developmental}, suggesting that learning and development play a key role in shaping our approximate numerosity representations. This perspective is further supported by computational modeling studies based on deep learning \citep{zorzi2018emergentist}, which have shown that a visual number sense can emerge as a higher-order statistical feature in multi-layer neural networks that learn to represent the structure of images with a varying number of items \citep{stoianov2012emergence,testolin2020visual}. These findings are aligned with the hypothesis that perceptual systems are adapted to the statistical properties of their surrounding environment \citep{fiser2010statistically}, which in deep learning models corresponds to the discovery of latent features that can be used to compactly encode the distribution of the training data \citep{hinton2007learning,zorzi2013modeling}.

A standard practice in numerosity perception studies has been to control the continuous properties of visual sets to minimize their correlation with numerosity \citep{piazza2010developmental}, while recent approaches systematically manipulate the stimulus space in order to quantitatively assess the influence of continuous magnitudes in numerosity judgments \citep{dewind2015modeling,dolfi2024measuring}. Developmental studies exploited these stimulus spaces to show that the encoding of numerical information is progressively sharpened,  whereas the influence of continuous cues decreases \citep{starr2017contributions,dolfi2024weaker}. However, non-numerical magnitudes can still bias numerosity judgments in adults, especially when the continuous properties are strongly incongruent with numerosity \citep{gebuis2012interplay}. Interestingly, deep neural networks show a similar interplay between numerical and non-numerical magnitudes \citep{testolin2020visual, dolfi2024weaker}.

Nevertheless, most studies of numerosity perception use computer-generated dot patterns as stimuli and include manipulations of stimulus properties that are unlikely to reflect their ``ecological'' distribution in natural environments, thereby introducing biases that could influence the way perceptual systems encode the relevant stimulus features. Similarly, most neural network models of numerosity perception have been trained on artificial stimuli that do not reproduce the statistical structure of our own developmental environment \citep{stoianov2012emergence,testolin2020numerosity,nasr2019number} For example, the deep learning model of \cite{stoianov2012emergence} was trained on a synthetic (i.e., computer-generated) dataset of images containing rectangular objects, where all numerosities appeared with the same frequency and cumulative area was orthogonally manipulated. Other studies investigated the representation of numerosity in convolutional neural networks  \citep{nasr2019number} trained for object recognition on Imagenet, a dataset of real images that mostly contain a single object on a non-cluttered background. Again, this is unlikely to  reflect the reality of the developmental environment that infants are stimulated with, and might tune onto.

Only one computational study has tried to exploit a more ecological training corpus, which contained synthetic images derived from real visual scenes featuring multiple objects \citep{testolin2020numerosity}. This allowed to investigate what could be the distribution of numerosity and some non-numerical magnitudes in more “naturalistic" contexts: it turned out that the size of individual items in the image was negatively correlated with numerosity, and the proportion of images containing a given number of items fell off very quickly as the number of items increased, following a Zipfian power law \citep{piantadosi2014zipf}.
However, in that study the original structure of the images was potentially altered during pre-processing stages (image resizing and de-overlapping of object bounding boxes), and the entire analysis relied on the availability of ground truth human annotations.

In this work, we propose an innovative computer vision pipeline that can be used to automatically process naturalistic images in order to identify and locate multiple objects and provide a precise segmentation of their silhouette. This method allows to estimate the distribution of numerosity in large-scale datasets depicting a variety of naturalistic environments, at the same time providing precise information related to the size of each individual item and its position in the visual scene.
This enables the investigation of the mutual correlations between numerosity and non-numerical visual magnitudes in naturalistic environments, even in the absence of human annotations. Extraction of such measures from large datasets is important from a theoretical and modeling perspective, but it also paves the way to more ecological investigations of numerosity perception in humans. Notably, findings based on artificial stimuli (dot patterns) might not necessarily generalize to object sets in natural visual scenes \citep{odic2023visual}, therefore a systematic and large-scale analysis of natural images allow us to establish to what extent the continuous dimensions co-vary with number, and to rank them in terms of the strength of this correlation.

The proposed pipeline relies on the combination of several state-of-the-art Artificial Intelligence (AI) techniques. The first processing stage exploits a multimodal large language model \citep{team2023gemini} to extract high-level semantic information from the raw pixels (i.e., the categories of the objects present in the image). This information is then used to guide an open-set object detector \citep{liu2023grounding}, which returns the exact location (bounding box) of each target object. Finally, a segmentation model \citep{kirillov2023segment} is used to extract the silhouette of the object within each bounding box, and thus measure its relative size.
Given the remarkable capabilities of multimodal large language models, one might wonder whether directly deploying one of these models could suffice to solve the task at hand (or, at least, whether it could serve as a strong baseline to evaluate our more sophisticated pipeline). Currently, the answer is no, since it has been thoroughly demonstrated that even the most advanced multimodal AI systems cannot reliably estimate the number of objects in a visual scene \citep{testolin2024visual}.

We calibrate and validate our pipeline using two large-scale datasets containing millions of photographs taken from a wide range of visual scenes \citep{everingham2010pascal,lin2014microsoftcoco} and show that it can detect and segment objects with an accuracy comparable to that of human annotators. This allows us to quantitatively characterize the statistical distribution of numerosity and non-numerical magnitudes in realistic image datasets, opening new research directions to study the statistical properties of even more ecological developmental environments.

\section{Materials and Methods}

\subsection{Naturalistic datasets}

\begin{figure}
    \centering
    \includegraphics[width=0.97\linewidth]{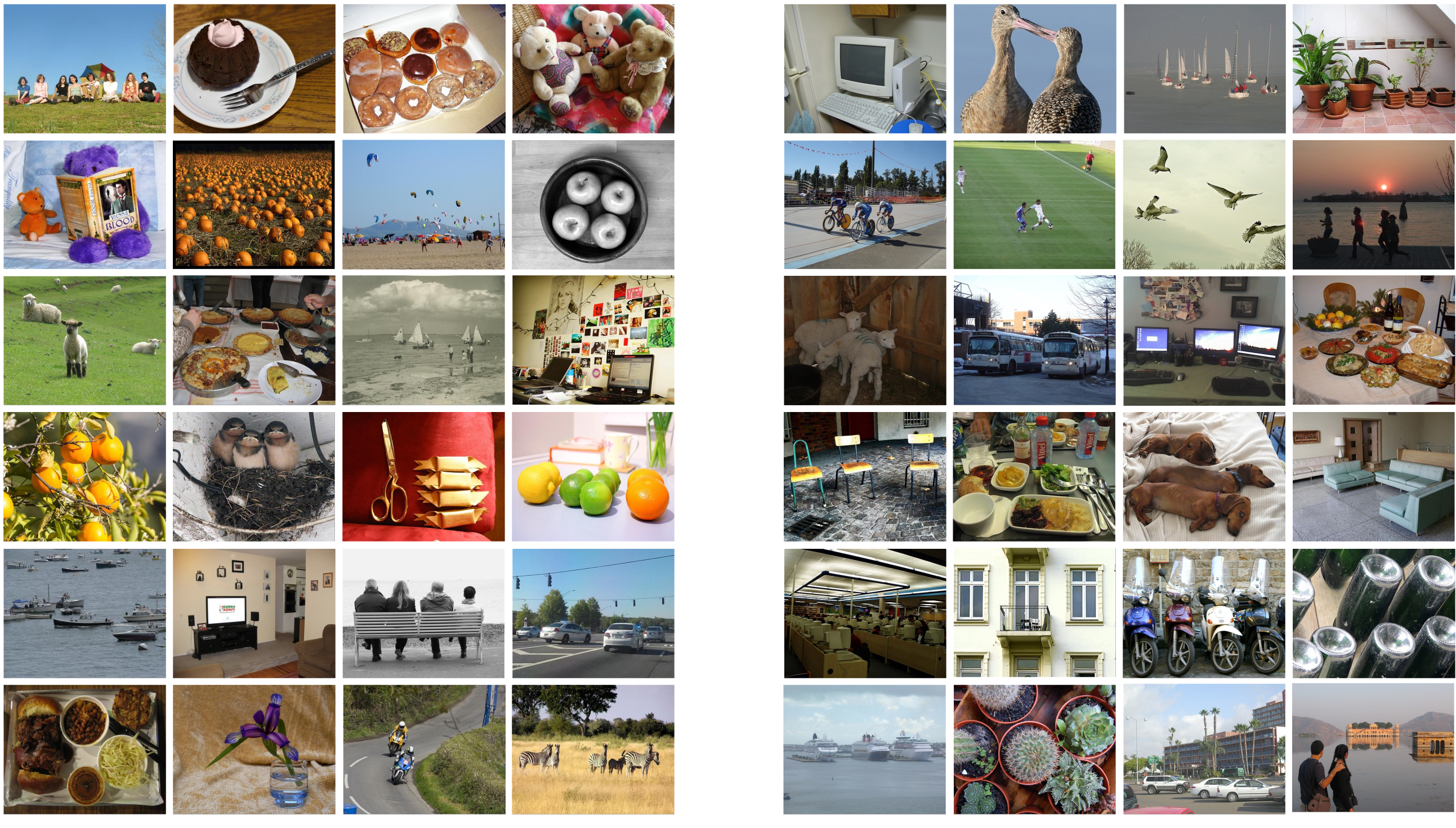}
    \caption{Representative images from MSCOCO (left) and PASCAL (right). Images in both datasets cover a wide range of themes, perspectives and object classes.}
    \label{fig:representative_images}
\end{figure}

As a first step, we surveyed the available literature to identify large-scale datasets of images that might constitute a good proxy for ecological environments. Considering to what extent the visual scenes in the dataset cohere with our daily life, the amount of images and object classes, the accuracy and detail of human annotations, and the availability of a clear labeling policy, we finally selected the MSCOCO \citep{lin2014microsoftcoco} and PASCAL \citep{everingham2010pascal} datasets for our investigation. We provide a brief overview of these two datasets below, while we refer the reader to the Supplementary Information for the description of other datasets that we did not eventually select for our analysis. Examples of images selected from these two datasets are shown in Figure \ref{fig:representative_images}.

We should note that our goal is to estimate the distribution of numerosity and non-numerical magnitudes in natural visual scenes, but we cannot argue that the images in selected datasets constitute a faithful representation of the sensory stimulation experienced by children or non-human animals during development. Indeed, all the available naturalistic datasets are made from photographs taken by human observers and may therefore contain aesthetic biases that do not necessarily reflect the structure of visual scenes perceived “in the wild". At the same time, however, we believe that the automatic analysis of large-scale realistic datasets constitutes the first step toward the development of innovative methods that can be used to analyze even more ecological stimuli, such as those containing egocentric videos recorded from the perspective of human infants \citep{sullivan2021saycam} or freely-behaving animals \citep{bar2024egopet}.

\subsubsection{Microsoft Common Objects in COntext (MSCOCO)}
Introduced in 2014, the MSCOCO dataset quickly became one of the reference datasets for the computer vision community. It is widely used for training deep learning models in tasks such as object detection, image classification, image segmentation, and image captioning. It contains hundreds of thousands of images enriched with detailed human annotations. The resolution of the images varies from $50\times50$ pixels to $640\times640$ pixels. The dataset encompasses 91 common object categories, balanced across multiple supercategories to ensure comprehensive coverage and practical applicability. Categories were selected through an integrative approach, combining sources such as PASCAL VOC, the most frequently used words denoting visually identifiable objects, and input from children naming objects in various environments. The supercategories include, among others: person, vehicle, outdoor, animal, accessory, sports, food, furniture, electronics, and appliances.
The final selection ensures each category's relevance and frequency in everyday scenes, with a significant number of instances per category to facilitate detailed and precise object localization.
Furthermore, the annotations distinguish between “things" and “stuff" categories. “Things" are discrete, \textit{countable} objects with clear boundaries, such as people, cars, and chairs, allowing for per-instance segmentation. In contrast, “stuff" categories, such as sky, grass, and water, represent amorphous, \textit{uncountable} regions without well-defined boundaries.
The size and variability of this dataset, combined with the rich and precise human annotations provided for each image, make it a perfect candidate for our investigation. We therefore decided to calibrate and validate our pipeline using this dataset.

\subsubsection{PASCAL Visual Object Classes (VOC)}
The PASCAL Visual Object Classes (VOC) dataset is another foundational resource in the field of computer vision. It is widely used to train deep learning models in tasks such as object detection, image classification, and image segmentation, and has previously been used to simulate the emergence of a visual number sense in neural network models \citep{testolin2020numerosity}. Its latest release includes 17,012 images featuring detailed annotations for 27,450 objects across 20 categories, such as vehicles, animals, and household items. Besides including object class labels, the human annotations of this dataset also provide bounding boxes for all objects in the visual field and, for a subset of 2913 images, also pixel-level segmentation of the object silhouette. These characteristics make it a good candidate for our investigation. We therefore decided to use this dataset as an additional validation resource for our pipeline.

\subsection{Automatic annotation pipeline}

Given that MSCOCO provides richer annotations for each image, we decided to calibrate our pipeline using this dataset, while PASCAL was used as a further independent validation of our analyses.

\begin{figure}
    \centering
    \includegraphics[width=0.9\linewidth]{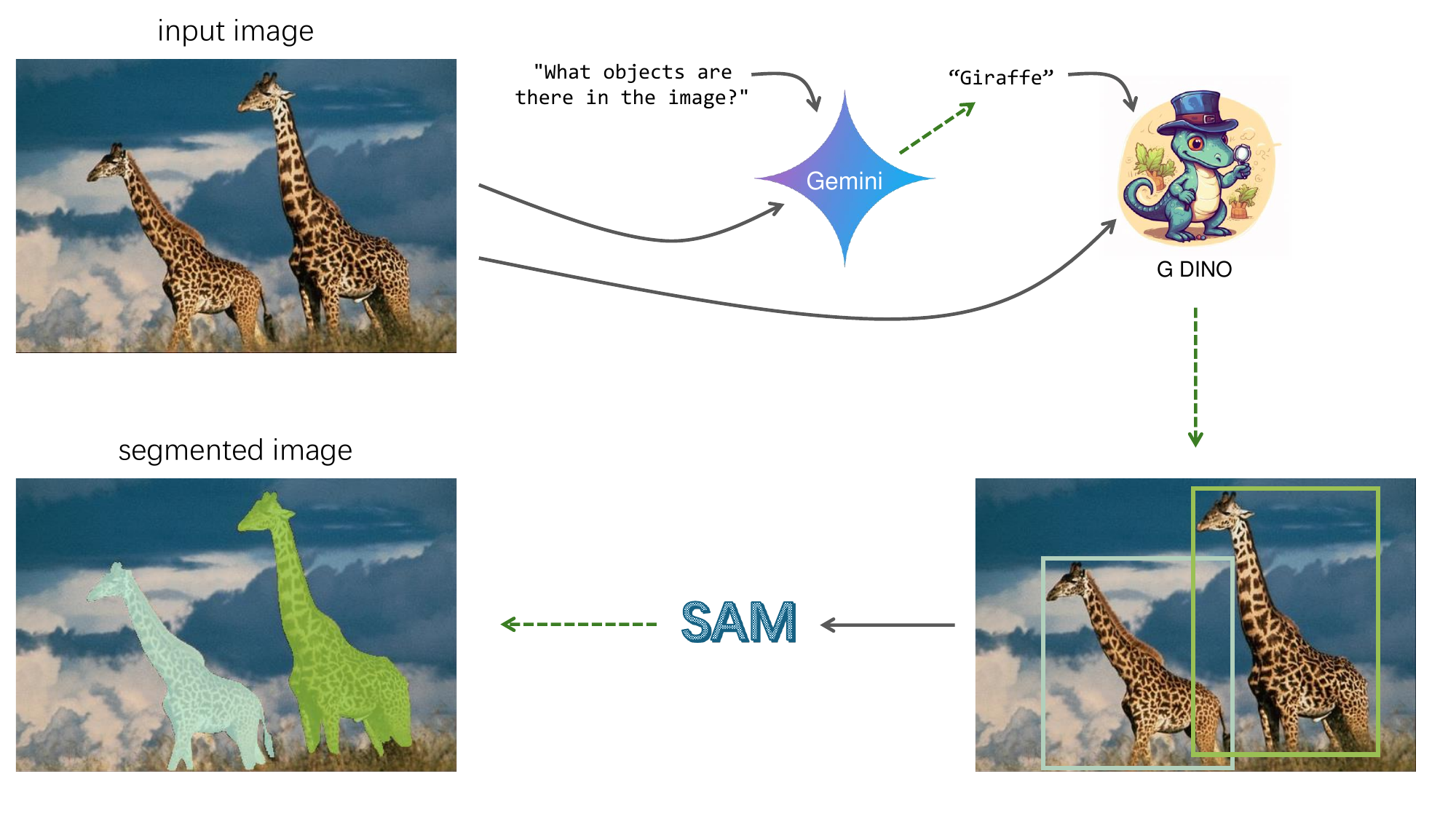}
    \caption{Schematic overview of our automatic annotation pipeline. Each image is initially provided as input to a multimodal large language model (Gemini), together with a textual prompt requesting to name the objects in the image. The object labels returned by Gemini and the original image are then given as input to an object detector model (Grounding DINO), which returns the coordinates of the object bounding boxes. The portion of the image corresponding to each bounding box is finally given as input to an image segmentation model (SAM), which returns the silhouette of each object.}
    \label{fig:pipe_illustartion}
\end{figure}

Our automatic annotation pipeline (see Figure \ref{fig:pipe_illustartion}) begins with presenting an image to a multimodal Large Language Model (LLM), which identifies and lists the names of potential objects present in the image. These objects are represented as textual labels. To locate them in the image, we then give the image and the labels returned by the LLM to an object detector model, which returns the coordinates of the bounding boxes for each identified object in the image. Finally, the coordinates are paired with the image and given to a segmentation model, which returns the segmentation mask (i.e., the silhouette) of each individual object. In the following, we provide the technical details of each processing stage.

\subsubsection{Object labeling}
We opted for using Google Gemini Pro Vision \citep{team2023gemini} as a generic multimodal LLM, as it achieves state-of-the-art performance on many visual reasoning benchmarks.
This powerful model was prompted by feeding each image paired with a structured textual prompt, with the goal of directing the LLM's attention on countable objects while ignoring backgrounds and continuous entities such as sky, lawn, and mountains. The specific prompt used was the following: \vspace{0.2cm} \\ 
\begin{small}
    \texttt{What objects/things are there in the image? Answer with only their singular names, and if there are more than one object categories, separate them with commas. If there are many objects of the same kind, name them only once. For example, for a scene with 3 apples and 2 bananas, you should answer: apple, banana. Focus on countable objects, ignore backgrounds and continuous stuff such as sky, lawn, mountains, etc. Add an identification code (i.e., 77592) before answering.}
\end{small}
\vspace{0.2cm}

Each LLM response was automatically verified by checking the presence of the identification code and making sure that the words were returned in the specified format indicated in the prompt. We then merged the LLM output with the list of object categories provided in the MSCOCO annotations. To this end, we first checked whether the words returned by the LLM were directly present in the MSCOCO list of categories, and discarded words belonging to the “stuff" (uncountable) categories.
If a word $\mathbf{A}$ did not appear in the list, we searched for the closest match by measuring its semantic similarity with each word $\mathbf{B}$ contained in the MSCOCO list, by creating linguistic embeddings for each word using the “word2vec-google-news-300" pre-trained model \citep{mikolov2013efficientword2vecmodel} and then measuring their cosine similarity:
\[
\text{cosine similarity} = \frac{\mathbf{A} \cdot \mathbf{B}}{\|\mathbf{A}\| \|\mathbf{B}\|}
\]
\noindent
If the highest similarity score was associated with a word from the list of “stuff" categories, it was removed from the final output, otherwise the original word was retained for the next processing stage.
This methodology ensured that each noun in the LLM response was accurately categorized or removed based on its semantic similarity to the predefined categories included in MSCOCO.

We should point out that the automatic LLM object labeling stage could also be skipped: one could directly provide a set of target categories to the next stage in the pipeline, which would allow, for example, to extract from images only a specific type of objects of interest (e.g., ``animals'' or ``manipulable objects'') or to focus on items with higher saliency.

\subsubsection{Object detection}
After acquiring the refined object labels, the pipeline uses the Grounding DINO object detection model \citep{liu2023grounding} to locate the target objects within the visual scene. Grounding DINO takes as input the image paired with the corresponding text labels and returns as output the coordinates of up to 900 object bounding boxes and their corresponding confidence scores. The pipeline initially filters the output by setting a relatively low threshold (i.e., 0.05) to keep as many boxes as possible before further processing. It then checks for strongly overlapping objects by calculating the Intersection of the Union (IoU) of all pairs of bounding boxes: if the IoU of two bounding boxes is greater than 0.95 (i.e., the objects are almost perfectly overlapping), it will only retain the one with a higher confidence score. Also bounding boxes occupying more than 95$\%$ of the image area were automatically removed. As a further step, the pipeline filters the bounding boxes again, this time setting a more conservative confidence threshold in order to only retain the most reliable detections. This threshold is optimized during the pipeline calibration stage, as explained in Sec. \ref{sec:pipeline_validation}.

\subsubsection{Object segmentation}
The final processing step deploys the Segment Anything Model (SAM) \citep{kirillov2023segment} to perform an accurate segmentation of the silhouette of each target object. SAM accepts various forms of input, such as point coordinates (center of mass) and bounding boxes, and returns the segmentation mask corresponding to the object of interest. The SAM architecture consists of three main components: an image encoder, a prompt encoder, and a mask decoder. The image encoder processes the entire image to create an embedding, which is a numerical representation of the image. The prompt encoder translates the textual user request into a format that the model can understand, integrating positional and textual information. Finally, the mask decoder uses this combined information to generate a segmentation mask, indicating the precise boundaries of each object.

\begin{figure}
\centering
\includegraphics[width=0.9\linewidth]{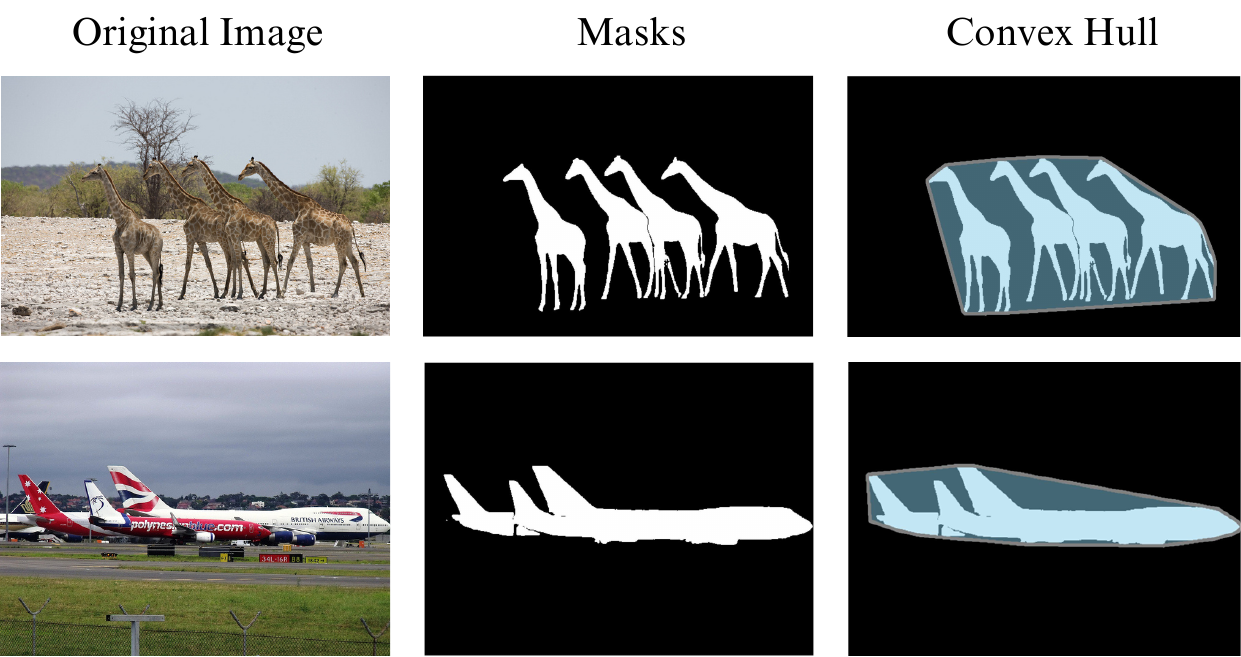}
\caption{Two examples depicting how occlusions are handled in the pipeline and how convex hull is considered (light blue area). To deal with object occlusions, the final silhouettes only contain the visible area of the segmented objects and the corresponding masks are treated separately in order to allow to properly count all objects in the image but at the same time allow to accurately estimate the size of their visible portion.}
\label{fig:occlusionhandle}
\end{figure}

We should note that our pipeline cannot estimate whether an object is occluded, nor the percentage of the object that is occluded.
In the case of partially occluded objects, however, the bounding box will always refer to the same object and therefore the numerosity estimate would not be impacted. The estimation of non-numerical magnitudes will also be unaffected, since we only consider the visible portions of each segmented object (see Figure \ref{fig:occlusionhandle}).

\subsubsection{Pipeline calibration} \label{sec:pipeline_validation}
The parameters of our pipeline (i.e., the confidence threshold used during the object detection stage) were calibrated with the goal of matching the distribution of numerosities derived from the MSCOCO human annotations. To do so, we implemented a grid-search procedure by systematically varying the value of the confidence threshold and measuring the similarity of the resulting numerosity distributions, using a Kolmogorov–Smirnov nonparametric test \citep{berger2014kolmogorov}. The threshold value ranged from \(0.10\) to \(0.45\), incremented by \(0.01\) at each step. This calibration step was crucial for determining the optimal parameter settings, ensuring that the distribution of numerosities returned from our automatic analysis was indeed comparable to that estimated from the human annotations.

\subsection{Extraction of numerosity and non-numerical magnitudes}

Numerosity simply corresponds to the number of bounding boxes returned by the Grounding DINO object detector. Besides numerosity, for each image we also estimated the following non-numerical magnitudes: cumulative area, per-item size, convex hull and density.
Cumulative area refers to the total number of pixels occupied by all segmented masks (i.e., object silhouettes). Per-item size describes the average number of pixels occupied by each segmented object. The convex hull is measured as the area of the convex polygon that includes all objects in the scene (see Figure \ref{fig:occlusionhandle}). Density is derived by dividing the numerosity by the convex hull.
As the size of the images varies even within the same dataset, the area occupied by the objects could be significantly affected by the image resolution: we therefore always considered relative magnitudes, by dividing the absolute value by the image size (i.e., the number of pixels in the image).

\section{Results}

\subsection{Pipeline validation}

\begin{figure}
    \centering
    \includegraphics[width=0.99\linewidth]{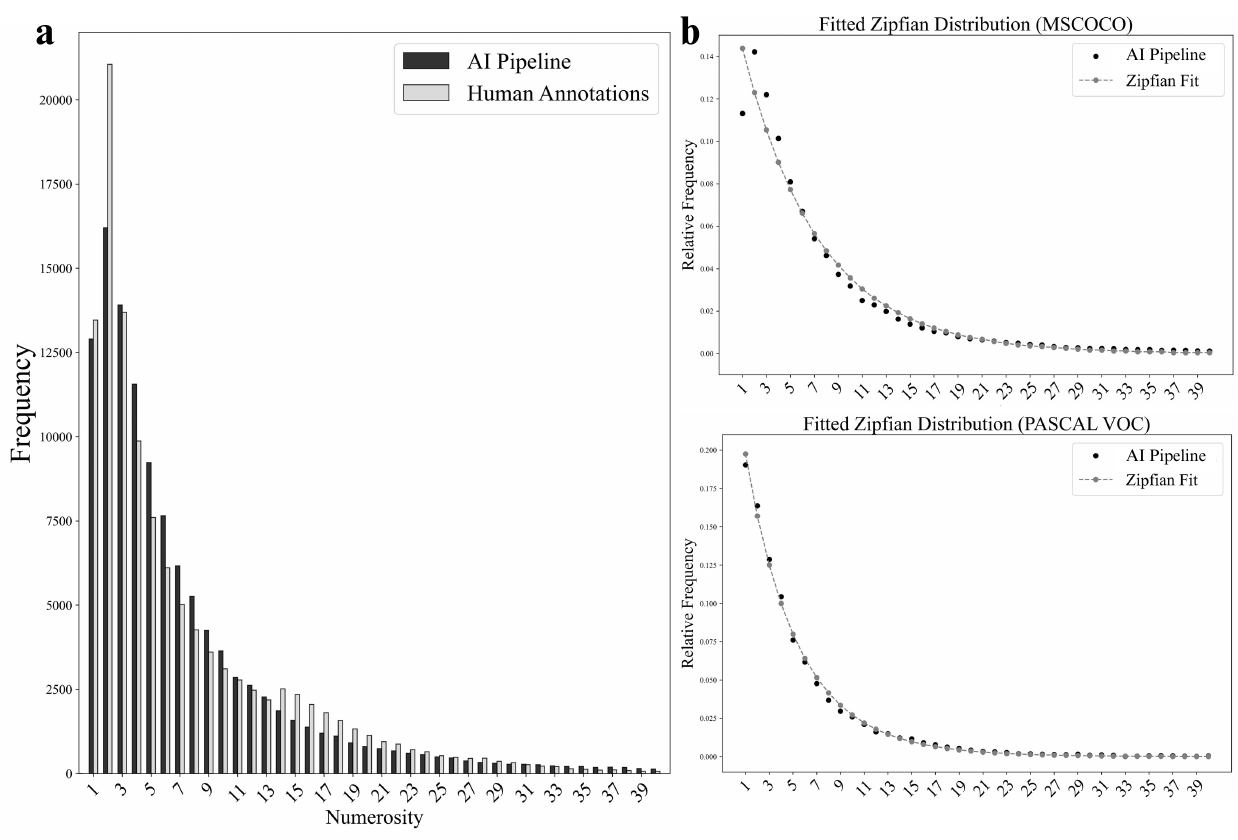}
    \caption{Distribution of numerosity in naturalistic visual scenes. a) Graphical comparison between the numerosity distribution computed over the MSCOCO human annotations and that derived from our automatic pipeline. b) Zipfian fits highlighting the power law decrease in numerosity frequency, observed for both the MSCOCO and the PASCAL natural datasets.}
    \label{fig:numerosity_distribution}
\end{figure}

\begin{figure}
    \centering
    \includegraphics[width=0.999\linewidth]{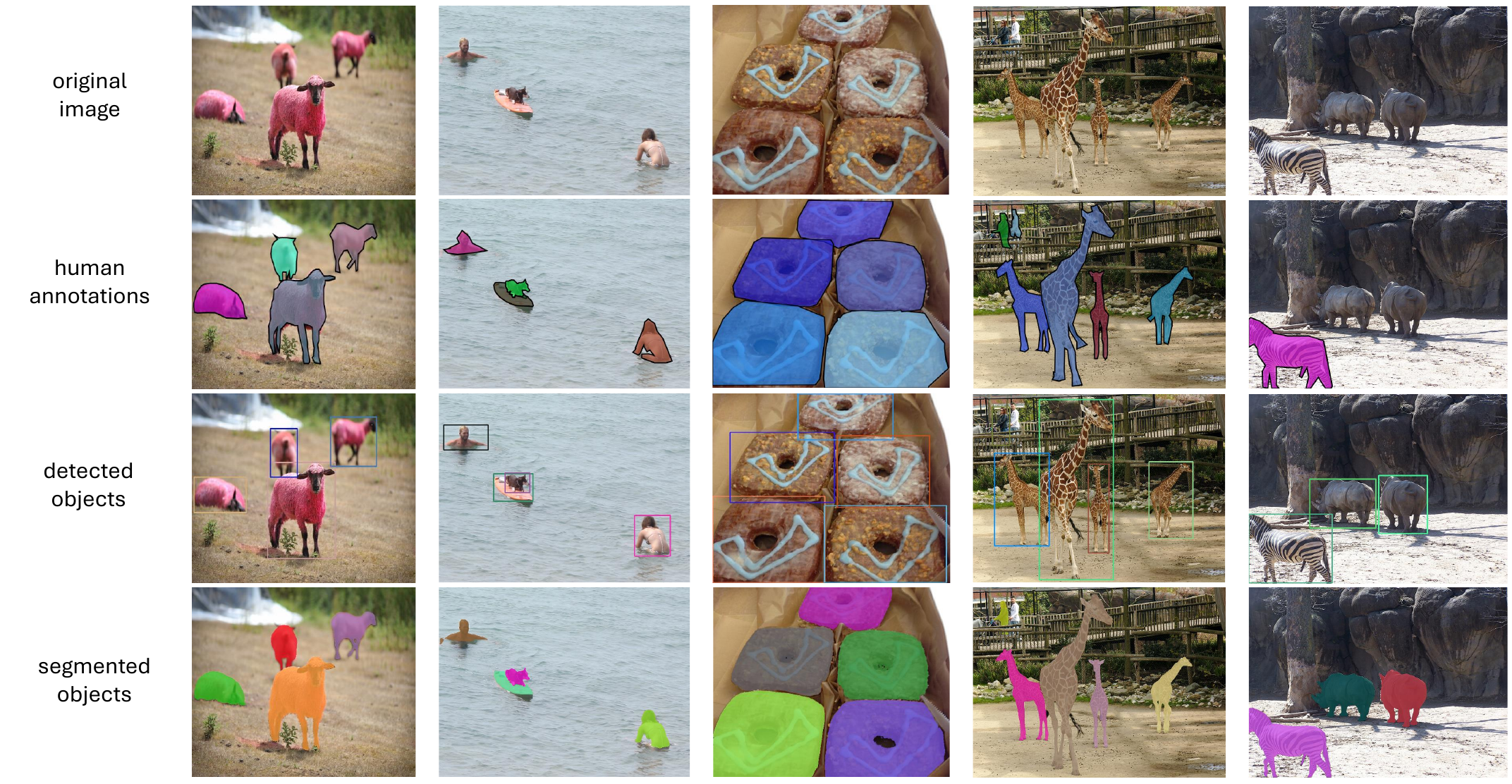}
    \caption{Examples of images comparing our automatic annotation pipeline with the ground truth human annotations. The first three columns represent good matches: our pipeline properly detects all tagged objects. The fourth column shows one case where our pipeline underestimates the numerosity of the visual set, since it neglects one of the two small people located in the upper-left portion of the visual field. The last column shows one case where our pipeline overestimates the number of objects with respect to the human annotations, but in fact the annotations are wrong (the two rhinos were not tagged).}
    \label{fig:over_under_estimation}
\end{figure}

As explained in Sec. \ref{sec:pipeline_validation}, we calibrated the pipeline parameters in order to closely match the distribution of numerosity derived from the ground truth human annotations. The grid search resulted in an optimal threshold value of $0.22$ for the object detector confidence score (lowest K-S statistics = $0.075$ with $p$-value $> 0.999$). The distributions of numerosity derived from our automatic pipeline and from the MSCOCO human annotations are shown in Figure \ref{fig:numerosity_distribution} (left panel). The mean absolute error between the ground truth numerosity and the value predicted by our pipeline was $3.76$. Together, these results suggest that, once calibrated, in most of the cases our pipeline can detect the correct number of objects in the images. Still, in some cases it can overestimate or underestimate the number of objects in the image with respect to the ground truth information provided by human annotators. However, one might wonder whether the human annotations should always be considered reliable, or whether in some cases our pipeline could be even more accurate than the putative ground truth: this indeed happens in some cases, as shown in Figure \ref{fig:over_under_estimation}.

\subsection{Distribution of numerosity and visual magnitudes}

As shown in Figure \ref{fig:numerosity_distribution} (left panel), the distribution of numerosities in natural visual scenes is strongly skewed, and small numerosities are overrepresented compared to larger ones. The highest number of items detected in a single image was 61, but in general images had much fewer items (only 0.08\% of the images had more than 40 items). This trend is well captured by a Zipfian power law, both for the MSCOCO dataset and for the PASCAL dataset (right panels in Figure \ref{fig:numerosity_distribution}). These findings are consistent with previous studies that highlighted a similar trend for the distribution of number words in linguistic corpora \citep{piantadosi2016rational}.

\begin{figure}
    \centering
    \includegraphics[width=0.99\linewidth]{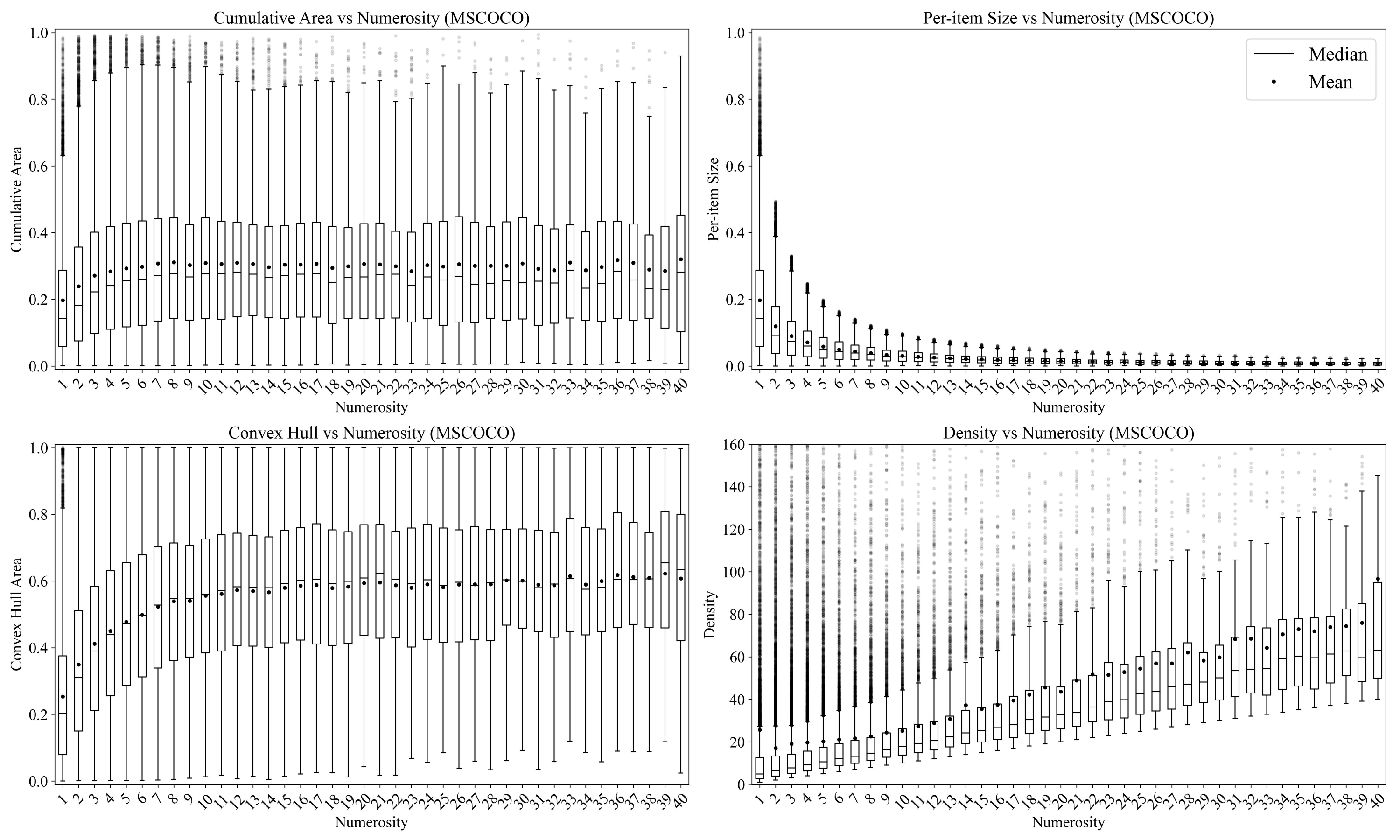}
    \caption{Box plots showing the relationship between numerosity and non-numerical visual magnitudes in the MSCOCO dataset. The top-left panel shows the correlation with relative cumulative area, the top-right panel shows the correlation with average the per-item size, the bottom-left panel shows the correlation with convex hull and bottom-right panel shows the correlation with display density.}
    \label{fig:COCO_correlations}
\end{figure}

\begin{figure}
    \centering
    \includegraphics[width=0.99\linewidth]{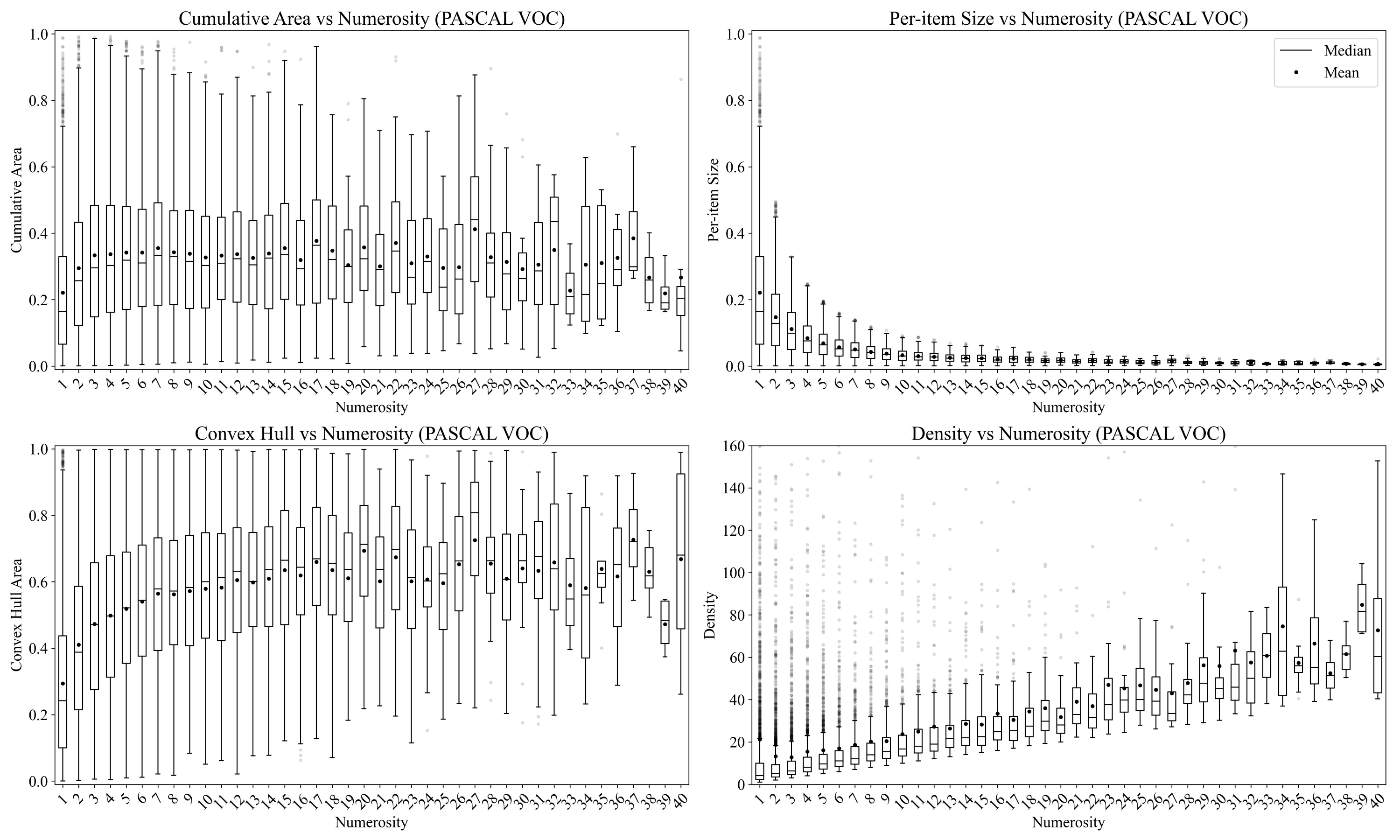}
    \caption{Box plots showing the relationship between numerosity and non-numerical visual magnitudes in the PASCAL dataset. The top-left panel shows the correlation with relative cumulative area, the top-right panel shows the correlation with average the per-item size, the bottom-left panel shows the correlation with convex hull and bottom-right panel shows the correlation with display density.}
    \label{fig:PASCAL_correlations}
\end{figure}

In Figure \ref{fig:COCO_correlations} and Figure \ref{fig:PASCAL_correlations} we show the relationship between numerosity and each non-numerical magnitude estimated by our pipeline in the MSCOCO and PASCAL datasets, respectively. We observe very similar trends in both datasets, suggesting that the naturalistic scenes represented in computer vision datasets have a common visual structure. The most correlated magnitudes are per-item size and density: When an image depicts several objects, they tend to appear smaller and cluster together, which is expected since the dimension of the visual field is constant. Interestingly, cumulative area does not appear to be strongly correlated with numerosity, suggesting that some of the synthetic datasets used to train deep learning models in previous studies \citep{stoianov2012emergence} embedded realistic distributions of non-numerical magnitudes. As expected, also the convex hull increases with numerosity, indicating that objects tend to spread in the visual field.

\subsection{Quantifying the correlations between numerosity and visual magnitudes}

We further quantified the relationship between numerosity and non-numerical magnitudes in terms of correlational structure within and across datasets. This analysis was carried out on images with numerosity outside the subitizing limit (n $>$ 4), since non-numerical magnitudes change drastically below this value. We first grouped together all images with the same numerosity, and then considered the average of the non-numerical magnitudes of each group to calculate the correlations. Holm correction was applied to account for multiple comparisons, and the correlation coefficients are represented as R values.

\begin{figure}
    \centering
    \includegraphics[width=0.85\linewidth]{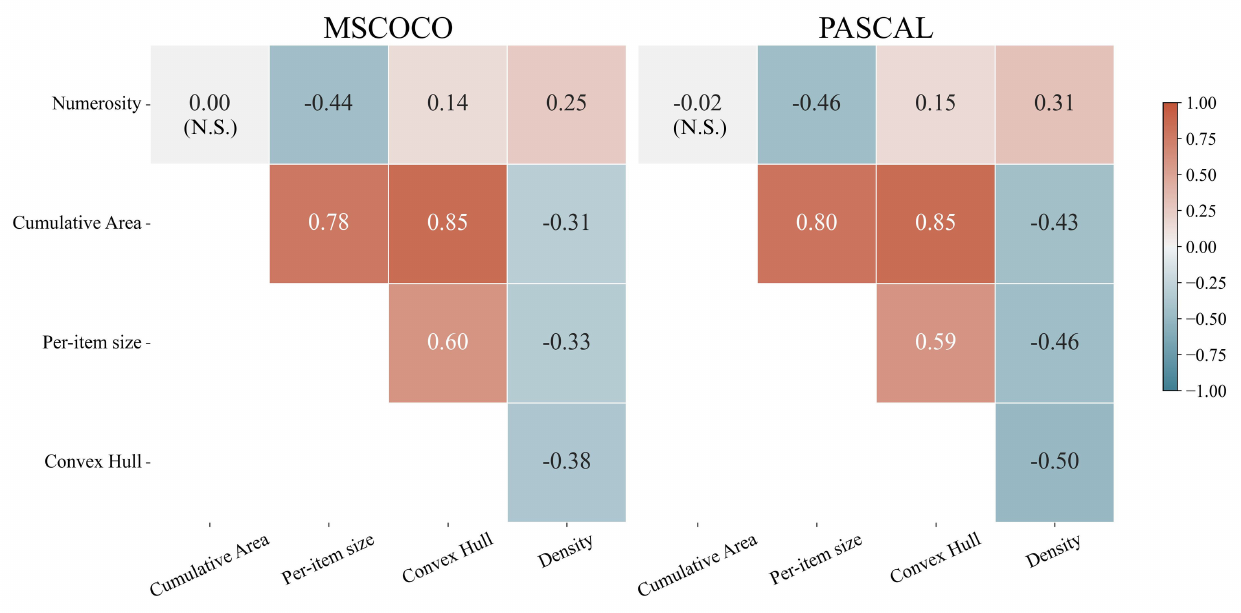}
    \caption{Correlation matrices between numerosity and non-numerical magnitudes in MSCOCO (left) and PASCAL (right). Color intensity corresponds to the strength of the correlation coefficient $R$, with warmer colors indicating positive correlations (all values significant with $p < 0.001$, except for those annotated as Not Significant [N.S.]).}
    \label{fig:corr_all}
\end{figure}

\begin{figure}
    \centering
    \includegraphics[width=0.85\linewidth]{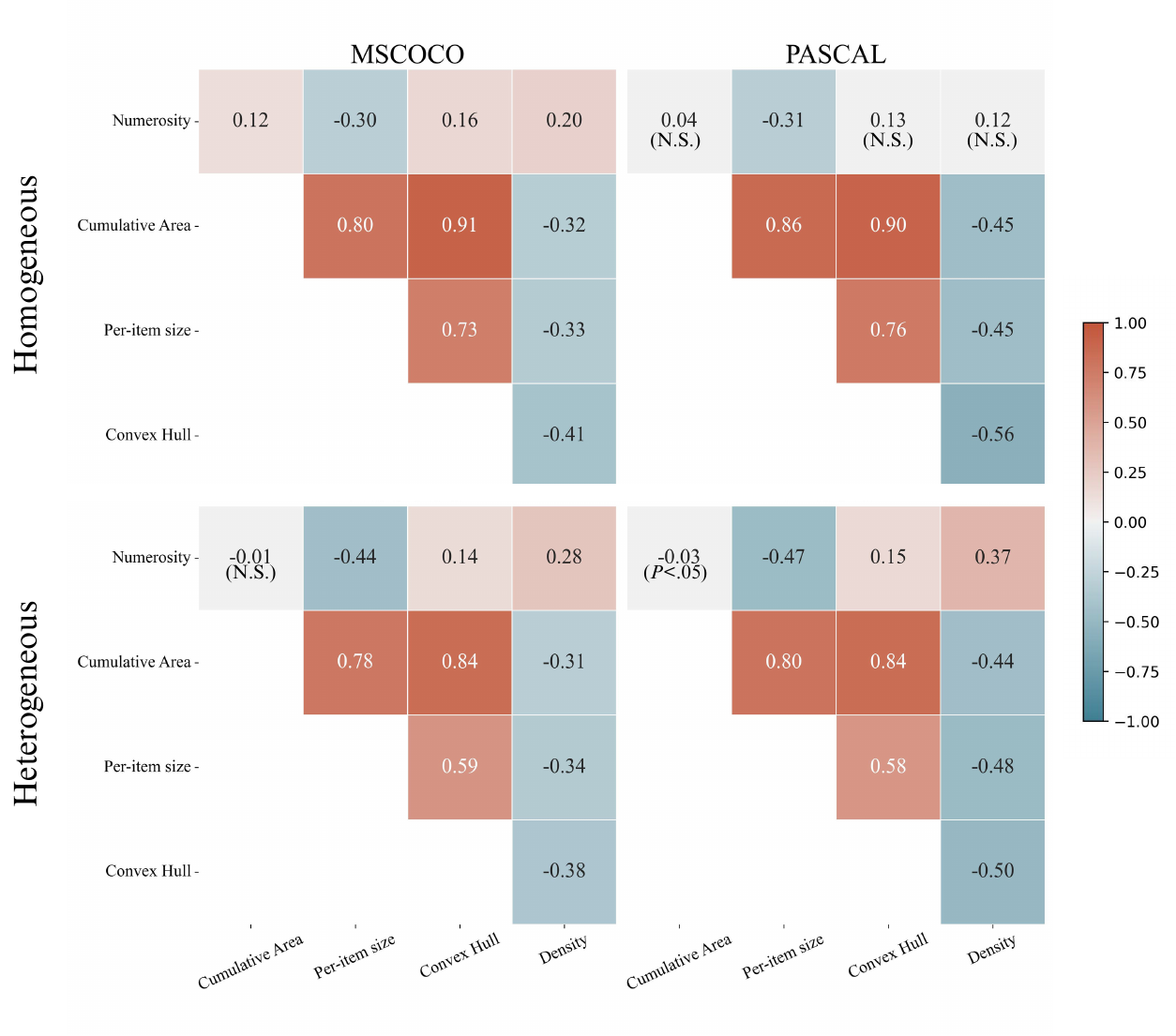}
    \caption{Correlation matrices between numerosity and all visual magnitudes in MSCOCO and PASCAL. The upper panels show the correlations for visual scenes featuring homogeneous object categories, while the lower panels show the correlations for heterogeneous visual scenes. Color intensity corresponds to the strength of the correlation coefficient $R$, with warmer colors indicating positive correlations (all values significant with $p < 0.001$, except for those annotated as Not Significant [N.S.] and the one annotated as $p < 0.05$).}
    \label{fig:corr_ann_mat}
\end{figure}

As shown in Figure \ref{fig:corr_all}, the correlation matrices for numerosity and non-numerical visual features are very consistent across the two datasets. We also measured the correlations separately for heterogeneous visual scenes (i.e., images featuring more than one object category) and homogeneous visual scenes (i.e., images annotated with only one type of objects), and found that the correlations remained very consistent (see Figure \ref{fig:corr_ann_mat}), indicating that the diversity of object categories within an image does not substantially alter the relationship between numerosity and non-numerical magnitudes. We systematically evaluated the consistency among different datasets (MSCOCO / PASCAL), different types of visual scenes (homogeneous / heterogeneous) and different sources of object labels (LLM / human annotations) by computing the correlation between the resulting correlation matrices. We found that results were always consistent (all pairwise correlations $> 0.977$).


\section{Discussion}

Overall, our results show that by tuning our image processing pipeline we could achieve a satisfactory match with the image annotations provided by human experts during the creation of the MSCOCO dataset.
This allowed us to automatically estimate the distribution of numerosity and non-numerical magnitudes in the MSCOCO dataset, but also in the PASCAL dataset, which does not contain such detailed human annotations.

In line with previous findings \citep{testolin2020numerosity}, it turned out that in both datasets there are many more images containing a few items, while larger numerosities are underrepresented. This trend is well captured by a Zipfian power law resembling the estimates of the frequency of occurrence of number words in linguistic corpora \citep{piantadosi2016rational}. Furthermore, our analyses revealed that the correlational structure for numerosity and continuous magnitudes is stable across two datasets as well as for homogeneous vs. heterogeneous scenes. 
Although many authors have previously postulated that changes in number are correlated with changes in other stimulus dimensions in natural environments (for discussion, see \cite{cantrell2013open}), our study provides the first large-scale quantitative characterization of the mutual correlation between numerosity, cumulative area, per-item area, convex hull and density.
This could explain why infants, at least sometimes, attend to and may even rely on a non-numerical dimension when comparing arrays that differ in number (e.g., \cite{clearfield1999number,feigenson2002infants}), and why a strong manipulation of some (but not all) continuous visual magnitudes affects numerosity judgements also in adults (e.g., \cite{gebuis2012interplay}).

Some authors have proposed that such interference effects occur because number is not perceived (or encoded) independently of other magnitudes \citep{lourenco2023theory}.
Others have argued that numerosity is, in fact, derived from the combination of different continuous sensory variables; that is, it is just the byproduct of a “general sense of quantity” \citep{leibovich2017sense}. However, our analyses suggest that non-numerical biases might occur because our visual system learns the natural correlations between number and non-numerical magnitudes in natural environments, and under certain circumstances such mutual correlations could be exploited to carry out numerosity judgements using other features as proxies: for example, density might be used to compare sets when density information is particularly salient or easier to extract than discrete quantity, such as for visual scenes with many objects \citep{anobile2015mechanisms}. Moreover, the influence of specific visual cues on numerosity judgements might be related to the degree of shift from the ecological distribution.

Although we calibrated and validated our pipeline using two large-scale datasets containing thousands of photographs depicting objects in a variety of natural environments, we cannot argue that this type of image faithfully represents the perceptual environment experienced by children and non-human animals during development.
For example, a recent study has shown that correlations between numerosity and continuous magnitudes are stronger in naturalistic photographs than in children’s counting book illustrations \citep{sanford2024non}, although it should be noted that in such case the sample of naturalistic images only contained visual scenes depicting birds, sheep, and motorcycles.

Still, we believe that our automated pipeline constitutes a valuable tool that could be used in future research to better characterize the statistical structure of even more ecological learning environments. For example, in future studies this method can be applied to automatically analyze longitudinal datasets of infant- and child-perspective egocentric videos \citep{sullivan2021saycam}, or recently published datasets containing egocentric videos of freely-behaving animals \citep{bar2024egopet}. Nevertheless, several refinements will be required in order to robustly deploy our pipeline in such challenging settings. Indeed, egocentric videos are normally recorded using low-resolution, wide-angle cameras, which might distort object appearance in the visual field and might thus require further processing stages. Furthermore, videos often contain blurred and noisy frames, which might not be correctly parsed by current computer vision models.

Another important limitation of the current pipeline is that it does not take object salience into account, while it is well-known that our enumeration skills can be strongly influenced by attention and saliency \citep{melcher2011role}. Despite recent improvements in deep learning models for salient object detection \citep{borji2019salient}, also in the case of video streams \citep{wang2019revisiting}, incorporating saliency information into our pipeline will require extensive experimentation and calibration.

\section{Conclusion}

In this work, we explored the use of modern Artificial Intelligence tools and large-scale machine-vision datasets to estimate the distribution of numerosity and non-numerical visual magnitudes in naturalistic scenes.
To this aim, we designed and implemented a computer vision pipeline that can be used to automatically process images, returning the list of objects contained in the picture along with detailed information about object locations and contours. Our method allowed us to characterize the distribution of numerosity and its putative correlations with a variety of continuous visual features such as cumulative area, item size, convex hull and density. We argue that considering the ecological pattern of covariance in natural visual scenes is important to understand the influence of non-numerical visual cues on numerosity judgements, and we believe that future studies could build on our approach to provide even more accurate estimates of the naturalistic distributions experienced by children during learning and development.

\vspace{0.5cm}

\bmhead{Authors Contributions}
All authors contributed to the conception of the study and the design of the processing pipeline. K.H. conducted the experiments and analyzed the results. A.T. and K.H. wrote the first draft of the manuscript. All authors revised the final version of the manuscript. 

\bmhead{Funding}
This project was partially supported by the Italian Ministry of Education and Research [PRIN Grant 2022EBC78W] and by the European Union - NextGenerationEU as part of the National Recovery and Resilience Plan (PNRR) [BAC FAIR SP10 J93C24000320007 GROUNDEEP]. K.H. acknowledges the support of the China Scholarship Council (ID: 202307820031).

\bmhead{Data and code availability}
The data and code used in this work are freely available online on \href{https://github.com/CCNL-UniPD/natural_scenes_pipeline}{GitHub}.

\bmhead{Competing Interests}
The authors declare no competing interests.

\bmhead{Supplementary information}
Supplementary data associated with this article can be found in the online version.

\bibliography{psychresearch}

\backmatter

\section*{Supplementary Information}

\subsection*{Other naturalistic datasets}

\subsubsection*{ImageNet}
ImageNet \citep{imagenet_cvpr09} is a large-scale image database released in 2009, organized according to the WordNet semantic hierarchy with a primary focus on nouns. This database has been progressively refined and nowadays contains more than 1 million of images representing more than 1000 object categories. ImageNet was created by the machine learning community to train object recognition systems, and indeed mostly contains images depicting a single item. The dataset is annotated only with labels representing the category of the objects in each image. Therefore, we cannot infer the position of the objects in the visual field or their size. For these reasons, we did not consider it as a good candidate for our investigation.

\subsubsection*{CityScapes}
The Cityscapes dataset \citep{Cordts2016Cityscapes} was designed to improve the semantic understanding of urban street scenes and includes several key features and annotations tailored to this end. It offers dense semantic and instance segmentation for vehicles and people across 30 classes from 50 cities. It was captured throughout spring, summer, and fall, primarily under good weather conditions. The dataset contains 5,000 finely annotated images and 20,000 with coarse annotations, each derived from video snippets that provide preceding and trailing frames. Additional metadata includes corresponding right stereo views, GPS coordinates, vehicle odometry, and environmental temperature. The labeling procedure was designed to maintain integrity in complex scenes where multiple classes or dynamic elements are present, such as vehicles and pedestrians. However, we believe that the narrow domain of this dataset does not make it a good candidate for our purposes.
 
\subsubsection*{Google Open Image}
The Google Open Images dataset \citep{openimages} is another vast resource, featuring approximately 9 million images annotated with labels, bounding boxes, segmentation masks, and visual relationships. The resolution of the images varies from about $50\times50$ pixels to around $4000\times4000$ pixels. The annotations include 16 million bounding boxes for 600 object classes and segmentation masks for 2.8 million object instances across 350 classes. This dataset was initially considered for our analysis; however, after some pre-processing, we found that this dataset, despite its abundant images and annotations, has several issues: the hierarchy of object categories is not clearly defined, and there are duplicated sub-categories and overlapping masks tagged using different labels (e.g., synonyms or supercategories). Furthermore, the ground truth information was derived in a semi-automatic fashion, introducing noise in the annotation process. For these reasons, we eventually decided to discard it.


\end{document}